\documentclass{llncs}
\usepackage[table]{xcolor}
\usepackage{amsmath}
\usepackage{graphicx}        
\usepackage{multicol}        
\usepackage{multirow}
\usepackage{lscape}
\usepackage{float}
\usepackage{url}
\usepackage{caption,subfig}

\def\dae{{\em Divide-and-Evolve}}
\def\DAE{{\sc DaE}}
\def\DAEX{{\sc DaE$_{\text{X}}$}}
\newcommand{\DAEYAHSP}{{\sc DaE$_{\text{YAHSP}}$}}
\def\PARADISEO{{\sc ParadisEO-MOEO}}
\def\YAHSP{{\sc YAHSP}}

\def\ZENO{{\sc Zeno}}
\def\MULTIZENO{{\sc MultiZeno}}
\def\PARAMILS{{\sc ParamILS}}

\begin{document}

\mainmatter              
%
\title{Multi-Objective AI Planning: \\ Evaluating \DAEYAHSP\ on a Tunable Benchmark}
%
\titlerunning{Evolutionary Multi-Objective AI Planning}  
%
\author{M.~R. Khouadjia \inst{1} \and M. Schoenauer\inst{1}\and
V. Vidal\inst{2}  \and J. Dr\'eo\inst{3} \and P. Sav\'eant\inst{3}}
%
\authorrunning{Mostepha~R. Khouadjia et \textit{al.}} 
%
\tocauthor{Mostepha~R. Khouadjia, Marc Schoenauer, Vincent Vidal, Johann Dr\'eo, and Pierre Sav\'eant}
\institute{TAO Project-team, INRIA Saclay \&  LRI, Universit\'e Paris-Sud, Orsay, France\\
\email{\{mostepha-redouane.khouadjia,marc.schoenauer\}@inria.fr},\\ 
\and
ONERA-DCSD, Toulouse, France\\
\email{Vincent.Vidal@onera.fr}\\
 \and
 THALES Research \& Technology, Palaiseau, France\\
 \email{\{johann.dreo, pierre.saveant\}@thalesgroup.com}\\
}

\maketitle              
\renewcommand{\thefootnote}{}
\begin{abstract}
All standard Artifical Intelligence (AI) planners to-date can only handle a single objective, and the only way for them to take into account multiple objectives is by aggregation of the objectives. Furthermore, and in deep contrast with the single objective case, there exists no benchmark problems on which to test the algorithms for multi-objective planning.

\dae\ (\DAE) is an evolutionary planner that won the (single-objective) deterministic temporal satisficing track in the last International Planning Competition. Even though it uses intensively the classical (and hence single-objective) planner \YAHSP\ ({\em Yet Another Heuristic Search Planner}), it is possible to turn \DAEYAHSP\ into a multi-objective evolutionary planner.

A tunable benchmark suite for multi-objective planning is first proposed, and the performances of several variants of multi-objective \DAEYAHSP\ are compared on different instances of this benchmark, hopefully paving the road to further multi-objective competitions in AI planning.\footnote{This work was partially funded by DESCARWIN ANR project (ANR-09-COSI-002).}


\end{abstract}
\section{Introduction}

An AI Planning problem (see e.g. \cite{AIplanningBook2004}) is defined by a set of predicates, a set of actions, an initial state and a goal state. A state is a set of non-exclusive instantiated predicates, or (Boolean) atoms. An action is defined by a set of {\em pre-conditions} and a set of {\em effects}: the action can be executed only if all pre-conditions are true in the current state, and after an action has been executed, the effects of the action modify the state: the system enters a new state.
A plan in AI Planning is a sequence of actions that transforms the initial state into the goal state. 
The goal of AI Planning is to find a plan that minimizes some quantity related to the actions: number of actions, or sum of action costs in case actions have different costs, or makespan in the case of temporal planning, when actions have a duration and can eventually be executed in parallel. All these problems are P-SPACE.

A simple planning problem in the domain of logistics is given in Figure \ref{fig.instance}: the problem involves cities, passengers, and planes. Passengers can be transported from one city to another, following the links on the figure. One plane can only carry one passenger at a time from one city to another, and the flight duration (number on the link) is the same whether or not the plane carries a passenger (this defines the {\em domain} of the problem). In the simplest non-trivial {\em instance} of such domain, there are 3 passengers and 2 planes. In the initial state, all passengers and planes are in {\tt city 0}, and in the goal state, all passengers must be in {\tt city 4}. The not-so-obvious optimal solution has a total makespan of 8 and is left as a teaser for the reader.

AI Planning is a very active field of research, as witnessed by the success of the ICAPS conferences (\url{http://icaps-conferences.org}), and its Intenational Planning Comptetition (IPC), where the best planners in the world compete on a set of problems. This competition has lead the researchers to design a common language to describe planning problems, PDDL (Planning Domain Definition Language). Two main categories of planners can be distinguished: {\em exact planners} are guaranteed to find the optimal solution \ldots if given enough time; {\em satisficing planners} give the best possible solution, but with no optimality guarantee. A complete description of the state-of-the-art planners is far beyond the scope of this paper. 

However, to the best of our knowledge, all existing planners are single objective (i.e. optimize one criterion, the number of actions, the cost, or makespan, depending on the type of problem), whereas most real-world problems are in fact multi-objective and involve several contradictory objectives that need to be optimized simultaneously. For instance, in logistics, the decision maker must generally find a trade-off between duration and cost (or/and risk). 

An obvious solution is to aggregate the different objectives into a single objective, generally a fixed linear combination of all objectives. Early work in that area used some twist in PDDL 2.0 \cite{do2003sapa,refanidis2003multiobjective,gerevini2008}. PDDL 3.0, on the other hand, explicitly offered hooks for several objectives x, and a new track of IPC was dedicated to aggregated multiple objectives: the ``net-benefit'' track took place in 2006 \cite{chen2006temporal} and 2008 \cite{edelkamp2009optimal}, \ldots but was canceled in 2011 because of the small number of entries.
In any case, no truly multi-objective approach to multi-objective planning has been proposed since the very preliminary proof-of-concept in the first \dae\ paper \cite{Schoenauer2006}. 

One goal of this paper is to build on this preliminary work, and to discuss various issues related to the challenge of solving multi-objective problems with an evolutionary algorithm that is heavily based on a single-objective planner (\YAHSP\ \cite{Vidal2004}) -- and in particular to compare different state-of-the-art multi-objective evolutionary schemes when used within \DAEYAHSP.
However, experimental comparison requires benchmark problems. Whereas the IPC have validated a large set of benchmark domains, with several instances of increasing complexity in each domain, nothing yet exists for multi-objective planning. The other goal of this paper is to propose a tunable set of benchmark instances, based on a simplified model of the IPC logistics domain \ZENO\ illustrated in Fig. \ref{fig.instance}. One advantage of this multi-objective benchmark is that the exact Pareto Front is known, at least for its simplest instances.
 
The paper is organized as follows: Section \ref{sec:dae} rapidly introduces \dae, more precisely the representation and variation operators that have been used in the single-objective version of \DAEYAHSP\ that won the temporal deterministic satisficing track at the last IPC in 2011. Section \ref{benchmark} details the proposed benchmark, called \MULTIZENO, and gives hints about how to generate instances of different complexities within this framework. Section \ref{sec:evolutionaryMOA} rapidly introduces the 4 variants of multi-objective schemes that will be experimentally compared  on some of the simplest instances of the \MULTIZENO\ benchmark and results of different series of experiments are discussed in Section \ref{sec:experiments}. Section \ref{sec:conclusion} concludes the paper, giving hints about further research directions.

\section{Divide-and-Evolve}
\label{sec:dae}
Let ${\cal P}_D(I,G)$ denote the planning problem defined on domain $D$ (the predicates, the objects, and the actions), with initial state $I$ and goal state $G$. In STRIPS representation model~\cite{Fikes1971}, a state is a list of Boolean atoms defined using the predicates of the domain, instantiated with the domain objects.  

In order to solve  ${\cal P}_D(I,G)$, the basic idea of \DAEX\ is to find a sequence of states $S_1, \ldots, S_n$, and to use some embedded planner $X$ to solve the series of planning problems ${\cal P}_D(S_{k},S_{k+1})$, for $k \in [0,n]$ (with the convention that $S_0 = I$ and $S_{n+1} = G$).
The generation and optimization of the sequence of states $(S_i)_{i \in [1,n]}$  is driven by an evolutionary algorithm. After each of the sub-problems ${\cal P}_D(S_{k},S_{k+1})$ has been solved by the embedded planner, the concatenation of the corresponding plans (possibly compressed to take into account possible parallelism in the case of temporal planning) is a solution of the initial problem. In case one sub-problem cannot be solved by the embedded solver, the individual is said {\em unfeasible} and its fitness is highly penalized in order to ensure that feasible individuals always have a better fitness than unfeasible ones, and are selected only when there are not enough feasible individual. A thorough description of \DAEX\ can be found in \cite{Bibai2010}. The following rest of this section will focus on the evolutionary parts of \DAEX.

\subsection{Representation and Initialization}
An individual in \DAEX\ is hence a  variable-length list of states of the given domain.
However, the size of the space of lists of complete states rapidly becomes untractable when the number of objects increases. Moreover, goals of planning problems need only to be defined as partial states, involving a subset of the objects, and the aim is to find a state such that all atoms of the goal state are true. An individual in \DAEX\ is thus a variable-length list of partial states, and a partial state is a variable-length list of atoms.

Previous work with \DAEX\ on different domains of planning problems from the
IPC benchmark series have demonstrated the need for a very careful choice of the atoms that are used to build the partial states \cite{bibai-EvoCOP2010}. 
The method that is used today to build the partial states is based on a heuristic estimation, for each atom, of the earliest time from which it can become true~\cite{Haslum2000}. 
These earliest start times are then used in order to restrict the candidate atoms for each partial state:
the number of states is uniformly drawn between 1 and the number of estimated start times; For every chosen time, the number of atoms per state is uniformly chosen between 1 and the number of atoms of the corresponding restriction.
Atoms are then added one by one: an atom is uniformly drawn in the allowed set of atoms (based on earliest possible start time), and added to the individual if it is not mutually exclusive (in short, {\em mutex}) with any other atom that is already there. Note that only an approximation of the complete mutex relation between atoms is known from the description of the problem, and the remaining mutexes will simply be gradually eliminated by selection, because they make the resulting individual unfeasible. 

To summarize, an individual in \DAEX\ is represented by a variable-length time-consistent sequence of partial states, and each partial state is a variable-length list of atoms that are not pairwise mutex. 

\subsection{Variation Operators}

Crossover and mutation operators are defined on the \DAEX\ representation in a straightforward manner - though constrained by the heuristic chronology and the partial mutex relation between atoms.

A simple one-point crossover is used, adapted to variable-length representation: both crossover points are independently chosen, uniformly in both parents. However, only one offspring is kept, the one that respects the approximate chronological constraint on the successive states. The crossover operator is applied with a population-level crossover probability.

Four different mutation operators are included: first, a population-level mutation probability is used; one an individual has been designated for mutation, the choice between the four mutation operators is made according to user-defined relative weights.
The four possible mutations operate either at the individual level, by adding (addState) or removing (delState) a state, or at the state level by adding (addAtom) or removing (delAtom) some atoms in a uniformly chose state. 

All mutation operators maintain the approximate chronology between the intermediate states (i.e., when adding a state, or an atom in a state), and the local consistency within all states (i.e. avoid pairwise mutexes).

\subsection{Hybridization}
\DAEX\ uses an external embedded planner to solve the sequence of sub-problems defined by the ordered list of partial states.
Any existing planner can in theory be used. However, there is no need for an optimality guarantee when solving the intermediate problems in order for \DAEX\ to obtain good quality results~\cite{Bibai2010}. Hence, and because several calls to this embedded planner are necessary for a single fitness evaluation, a sub-optimal but fast planner is used: \YAHSP~\cite{Vidal2004} is a lookahead 
strategy planning system for sub-optimal planning which uses the  actions in the relaxed plan to compute reachable states in order to speed up the search process.

For any given $k$, if the chosen embedded planner succeeds in solving $ P_{D} (S_k, S_{k+1} )$, the final complete state is computed by executing the solution plan
from $S_k$, and becomes the initial state of the next problem. If all the sub-problems are solved by the  embedded planner, 
the individual is called \textit{feasible}, and the concatenation of the plans for all sub-problems  is a
global solution plan for $P_{D} (S_{0} = I, S_{n+1} = G)$. However, this plan can in general be further optimized by rescheduling some of its actions, in a step called
compression. The computation of all objective values is done from the compressed plan of the given individual.
Finally, because the rationale for \DAEX\ is that all sub-problems should hopefully be easier than the initial global problem, and for computational performance reason, the search capabilities of the embedded planner \YAHSP\ are limited by setting a maximal number of nodes that it is allowed to expand to solve any of the sub-problems (see again \cite{Bibai2010} for more details).

\section{Multi-Objective Divide-and-Evolve}
\label{modae}
In some sense, the multi-objectivization of \DAEX\ is straightforward -- as it is for most evolutionary algorithms.
The ``only'' parts of the algorithm that require some modification are the selection parts, be it the parental selection, that chooses which individual from the population are allowed to breed, and the environmental selection (aka replacement), 
that decides which individuals among parents and offspring will survive to the next generation. 
Several schemes have been proposed in the EMOA literature (see e.g. Section \ref{sec:evolutionaryMOA}), and the end of this Section will briefly introduce the ones that have been used in this work. However, a prerequisite is that all objectives are evaluated for all potential solutions, and the challenge here is that the embedded planner \YAHSP\ performs its search based on only one objective.
 
\subsection{Multi-objectivization Strategies}
\label{sec:strategies}
Even though \YAHSP\  (like all known planners to-date) only solves planning problems based on one objective. However, it is possible since PDDL 3.0 to add some other quantities (aka Soft Constraints or Preferences \cite{gerevini2006preferences}) that are simply computed throughout the execution of the final plan, without interfering with the search. 

The very first proof-of-concept of multi-objective \DAEX\ \cite{Schoenauer2006}, though using an exact planner in lieu of the satisficing planner \YAHSP, implemented the simplest idea with respect to the second objective: ignore it (though computing its value for all individuals) at the level of the embedded planner, and let the evolutionary multi-objective take care of it. However, though \YAHSP\ can only handle one objective at a time, it can handle either one in turn, provided they are both defined in the PDDL domain definition file. Hence a whole bunch of smarter strategies become possible, depending on which objective \YAHSP\ is asked to optimize every time it runs on a sub-problem. Beyond the fixed strategies, in which \YAHSP\ always uses the same objective throughout  \DAEYAHSP\ runs, a simple dynamic randomized strategy has been used in this work: 
Once the planner is called for a given individual, the choice of which strategy to apply is made according to roulette-wheel selection based on user-defined relative weights; In the end, it will return the values of both objectives. 
It is hoped that the evolutionary algorithm will find a sequential partitioning of the problem that will nevertheless allow the global minimization of both objectives. Section \ref{resultsStrategies} will experimentally compare the fixed strategies and the dynamic randomized strategy where the objective that \YAHSP\ uses is chosen with equal probability among both objectives.

Other possible strategies include adaptive strategies, where each individual, or even each intermediate state in every individual, would carry a strategy parameter telling \YAHSP\ which strategy to use -- and this strategy parameter would be subject to mutation, too. This is left for further work. 

\subsection{Evolutionary Multi-Objective Schemes}
\label{sec:evolutionaryMOA}

Several Multi-Objective EAs (MOEAs) have been proposed in the recent years, and this work is concerned with comparing some of the most popular ones when used within the multi-objective version of \DAEYAHSP.
More precisely, the following selection/reproduction schemescan be applied to any representation, and will be experimented with here: NSGA-II~\cite{Deb2002}, SPEA2~\cite{Zitzler2002}, and IBEA~\cite{Zitzler2004}. They will now be quickly introduced in turn.


The {\bf Non-dominated Sorting Genetic Algorithm} (NSGA-II) has been proposed by Deb et \textit{al.}~\cite{Deb2002}. 
At each generation, the solutions contained in the current  population are ranked into successive Pareto fronts in the objective space. Individuals mapping to vectors from the first front all belong to
the best efficient set; individuals mapping to vectors from the second front all belong to the second best efficient set; and so on.
Two values are then assigned for every solution of the population. The first one corresponds to the rank of the Pareto front the corresponding solution
belongs to, and represents the quality of the solution in terms of convergence. The second one, the crowding distance, consists in
estimating the density of solutions surrounding a particular point in the objective space, and represents the quality of the solution in
terms of diversity.  A solution is said to be better than another solution if it has a better rank value, or in case of equality, if it has a larger crowding distance.

The {\bf Strength Pareto Evolutionary Algorithm} (SPEA)~\cite{Zitzler2001}, introduces an improved fitness assignment strategy. It intrinsically handles an internal fixed-size archive that is used during the selection step to create offspring solutions. At a given iteration of the algorithm, each population and archive member $x$ is assigned a strength value $S(x)$ representing the number
of solutions it dominates. Then, the fitness value $F (x)$ of solution $x$ is calculated by summing the strength values of all individuals that $x$ currently dominates. Additionally,
a diversity preservation strategy is used, based on a nearest neighbor technique.
The selection step consists of a binary tournament with replacement applied on the internal archive only.
Last, given that the SPEA2 archive has a fixed size storage capacity, a pruning mechanism based on fitness and diversity information is used when the non-dominated set is too large. 

The {\bf Indicator-Based Evolutionary Algorithm} (IBEA) \cite{Zitzler2004}
introduces a total order between solutions by means of a binary quality indicator. 
The fitness assignment scheme of this evolutionary algorithm is based on a pairwise comparison of solutions contained 
in  the current population with respect to a binary quality indicator $I$. Each individual $x$ is assigned a fitness value $F (x)$ measuring the ``loss in quality'' that would result from removing $x$ from the current
population. Different indicators can be used. The most two popular, that will be used in this work, are the additive $\epsilon$-indicator ($I_{\epsilon^+}$ ) and the hypervolume
difference indicator ($I_{H^-}$ )  as defined in ~\cite{Zitzler2004}. 
Each indicator  $I (x, x')$ gives the minimum value by which a solution $x \in X$  can be translated in the objective space to weakly dominate
another solution $x' \in X$. 
An archive stores solutions mapping to potentially non-dominated points in order to prevent their loss during the stochastic search process.

\begin{figure}[tb]
\begin{center}
 \includegraphics[width=0.5\textwidth]{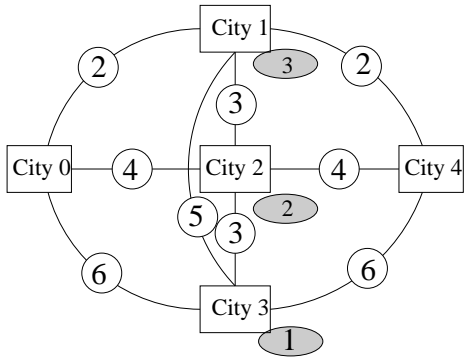}
\caption{A schematic view of \MULTIZENO, a simple benchmark transportation problem: Durations of available flights are attached to the corresponding edges, costs/risks are attached to landing in the central cities (in grey circles).}
\label{fig.instance}
\end{center}
\end{figure}

\section{A Benchmark Suite for Multi-Objective Temporal Planning}
\label{benchmark}

This section details the proposed benchmark test suite for multi-objective temporal planning, based on  the simple domain that is schematically described in Figure \ref{fig.instance}. The reader will have by now solved the little puzzle set in the Introduction, and found the solution with makespan 8 (flying 2 passengers to {\tt city 1}, one plane continues with its passenger to {\tt city 4} while the other plane flies back empty to {\tt city 0}, the plane in city {\tt city 4} returns empty to {\tt city 1} while the other plane brings the last passenger there, and the goal is reached after both planes bring both remaining passengers to {\tt city 4}). The rationale for this solution is that no plane ever stays idle.

In order to turn this problem into a not-too-unrealistic logistics multi-objective problem, some costs or some risks are added to all 3 central cities (1 to 3). This leads to two types of problems: In the \MULTIZENO$_{Cost}$, the second objective is an additive objective: each plane has to pay the corresponding tax every time it lands in that city; In the \MULTIZENO$_{Risk}$, the second objective is similar to a risk, and the maximal value encountered during the complete execution of a plan is to be minimized. 

In both cases, there are 3 obvious points that belong to the Pareto Front: the solution with minimal makespan described above, and the similar solutions that use respectively {\tt city 2} and {\tt city 3} in lieu of {\tt city 1}. The values of the makespans are respectively 8, 16 and 24, and the values of the costs are, for each solution, 4 times the value of the single landing tax, and exactly the value of the involved risk. For the risk case, there is no other point on the Pareto Front, as a single landing on a high-risk city sets the risk of the whole plan to a high risk. For the cost model however, there are other points on the Pareto Front, as different cities can be used for the different passengers. For instance, in the case of Figure \ref{fig.instance}, this leads to a Pareto Front made of 5 points, (8,12), (16,8), and (24,4) (going only through {\tt city 1}, {\tt 2} and {\tt 3} respectively), plus (12,10) and (20,6). Only the first 3 are the Pareto Front in the risk case.

\subsection{Tuning the Complexity}
There are several ways to make this first simple instance more or less complex. A first possibility is to  add passengers. In this work, only bunches of 3 passengers have been considered, in order to be able to easily derive some obvious Pareto-optimal solutions, using several times the little trick to avoid leaving any plane idle. For instance, it is easy to derive all the Pareto solutions for 6 and 9 passengers -- and in the following, the corresponding instances will be termed \MULTIZENO3,  \MULTIZENO6, and  \MULTIZENO9\ respectively (sub-scripted with the type of second objective -- cost or risk).

Of course, the number of planes could also be increased, though the number of passengers needs to remain larger than the number of planes to allow for non-trivial Pareto front. However, departing from the 3 passengers to 2 planes ratio would make the Pareto front not easy to identify any more.

Another possibility is to increase the number of central cities: this creates more points on the Pareto front, using either plans in which a single city is used for all passengers, or plans that use several different cities for different passengers (while nevertheless using the same trick to ensure no plane stays idle). In such configuration too the exact Pareto front remains easy to identify: further work will investigate this line of complexification.

%
%

\subsection{Modifying the shape of the Pareto Front}

\begin{figure*}[tb]
\centering{
\subfloat[$\alpha=1.1$]{ \includegraphics[width=0.32\textwidth]{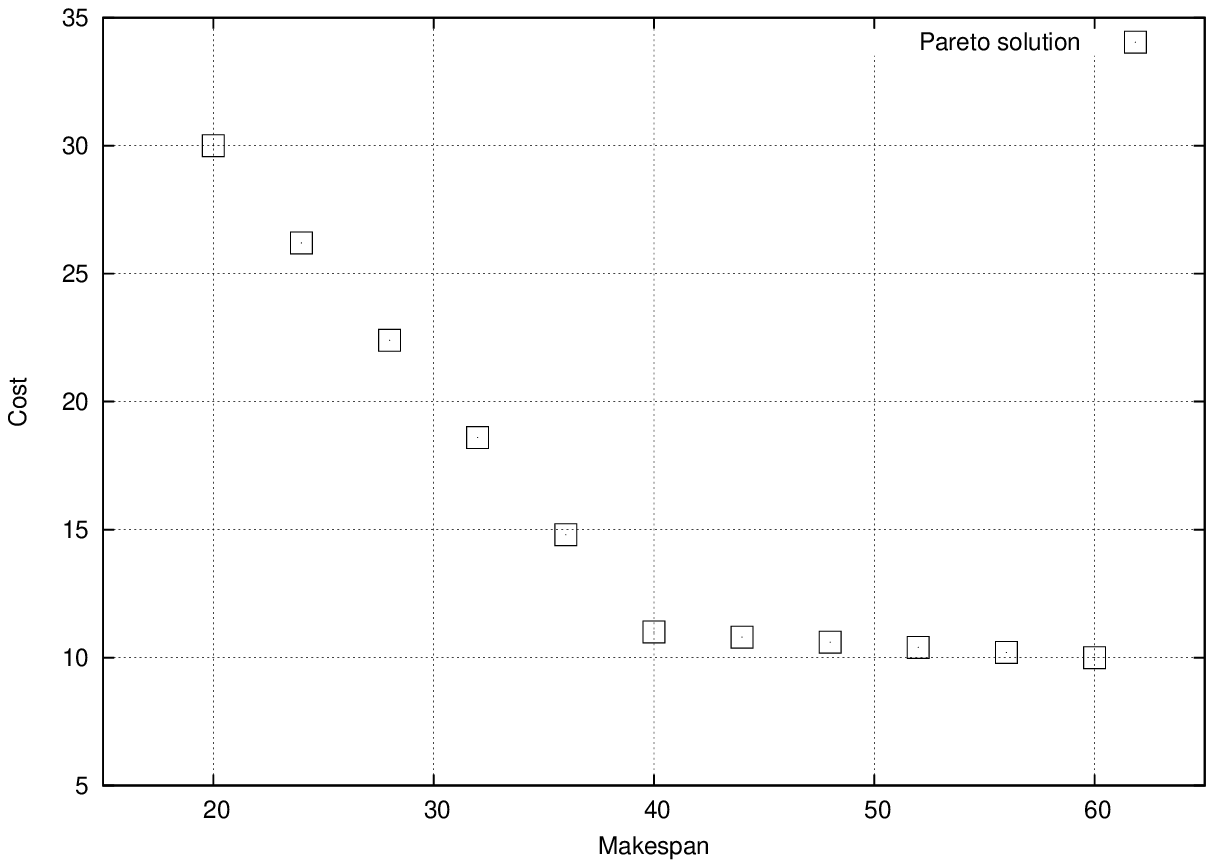}
\label{fig:zeno6i_Add}}
\subfloat[$\alpha=2$ (Fig. \ref{fig.instance})] { \includegraphics[width=0.32\textwidth]{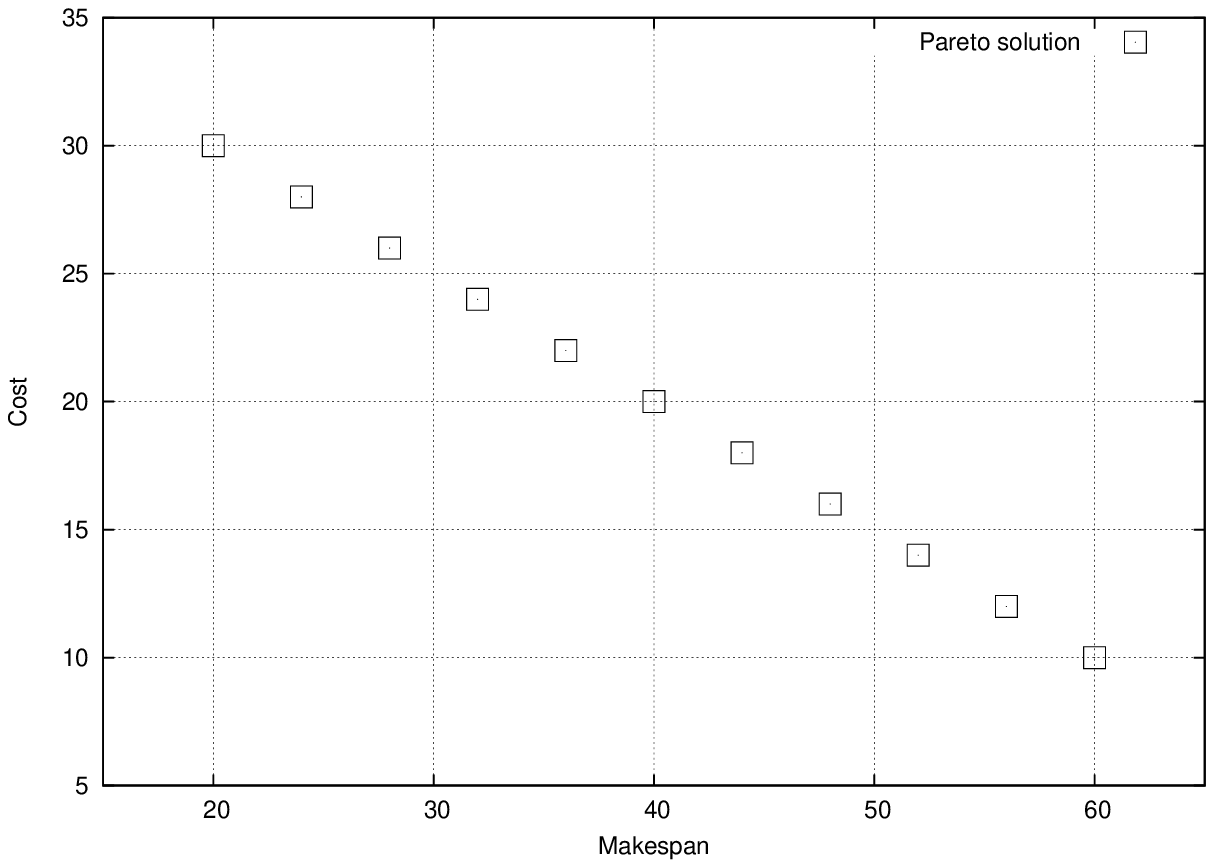}
\label{fig:zeno6e_Add}} 
\subfloat[$\alpha=2.9$] { \includegraphics[width=0.32\textwidth]{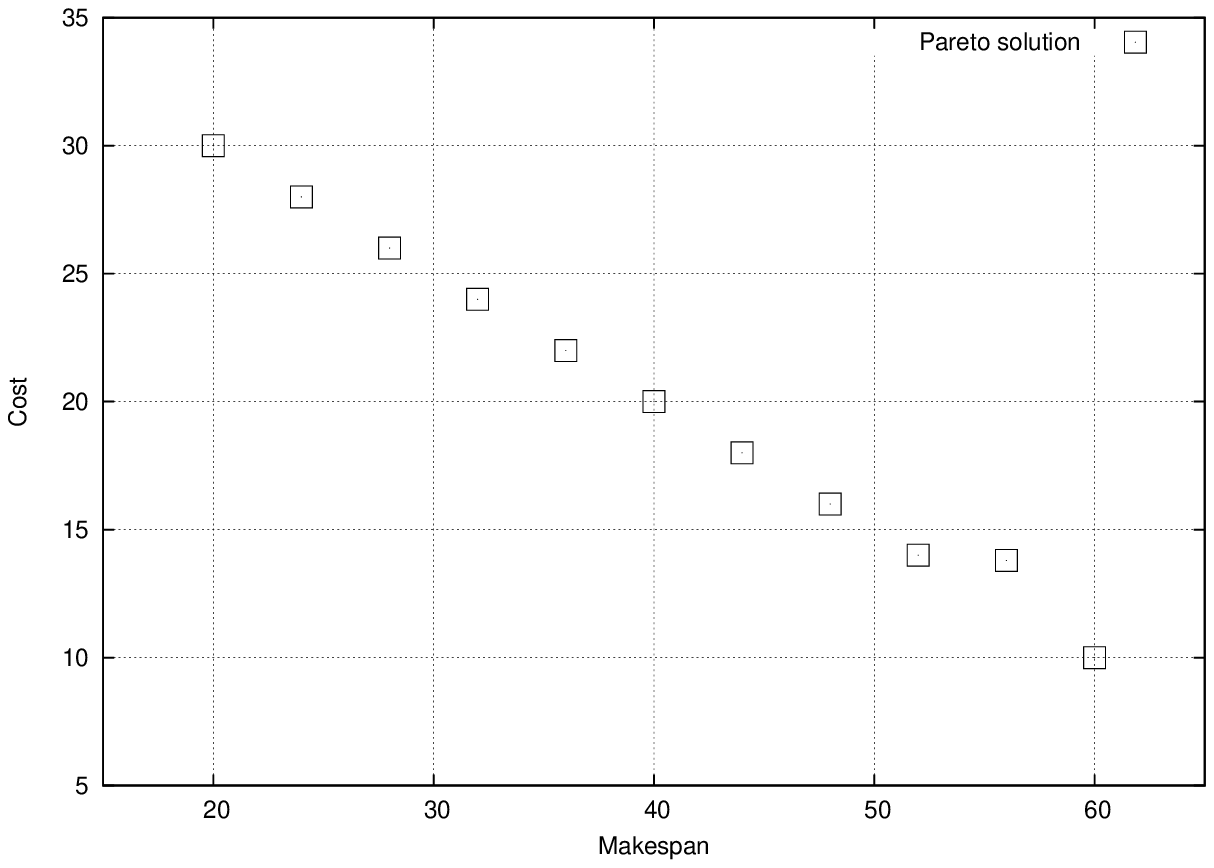}
\label{fig:zeno6s_Add}}\\
 
\caption{The exact Pareto Fronts for the \MULTIZENO6 problem for different values of the cost $\alpha$ of {\tt city2} (those of {\tt city1} and {\tt city3} being $3$ and $1$ respectively).}
\label{fig:zeno3ParetoFronts}}
\end{figure*}

Another way to change the difficulty of the problem without increasing its complexity is to tune the different values of the flight times and the cost/risk at each city. Such changes does not modify the number of points on the Pareto Front, but does change its shape in the objective space. For instance, simply modifying the cost $\alpha$ of {\tt city2}, the central city in Figure \ref{fig.instance},
between 1 and 3 (the costs of respectively {\tt city1} and {\tt city3}), the Pareto Front, which is linear for $\alpha=2$ becomes strictly convex for $\alpha < 2$ and strictly concave for $\alpha > 2$, as can be seen for two extreme cases ($\alpha = 1.1$ and $\alpha = 2.9$) on Figure \ref{fig:zeno3ParetoFronts}. Further work will address the identification of the correct domain parameters in order to reach a given shape of the Pareto front.

\section{Experimental Conditions}
\label{sec:condition}
\paragraph{Implementation:} All proposed multi-objective approaches (see Section \ref{sec:evolutionaryMOA}) have been implemented within the \PARADISEO\ framework \cite{paradiseo}. 
All experiments were performed on the \MULTIZENO3,  \MULTIZENO6, and  \MULTIZENO9 instances. The first objective is the makespan, and the second objective either the (additive) cost or the (maximal) risk, as discussed in Section \ref{benchmark}. The values of the different flight durations and cost/risks are those given on Figure \ref{fig.instance} except otherwise stated.

\paragraph{Parameter tuning:} All user-defined parameters have been tuned using the  framework \PARAMILS\
\cite{ParamILS-JAIR}.  \PARAMILS\ handles any parameterized algorithm whose parameters can be discretized. Based on Iterated Local Search (ILS), \PARAMILS\ searches
through the space of possible parameter configurations, evaluating configurations by running the algorithm to be optimized on a set of benchmark instances, searching for the configuration that yields
overall best performance across the benchmark problems. Here, both the parameters of the multi-objective algorithms (including the internal parameters of the variation operators -- see \cite{Bibai:2010:GPT:1830483.1830528}) and \YAHSP\ specific parameters (including the relative weights of the possible strategies (see Section \ref{sec:strategies}) have been subject to \PARAMILS\ optimization. For the purpose of this work, parameters were tuned anew for each instance (see \cite{Bibai:2010:GPT:1830483.1830528} for a discussion about the generality of such parameter tuning, that falls beyond the scope of this paper).

\paragraph{Performance Metric:} The quality measure used by \PARAMILS\ to optimize \DAEYAHSP\ is the unary hypervolume  $I_{H^-}$~\cite{Zitzler2004} of the set of non-dominated points output by the algorithm with respect to the complete true Pareto front (only instances where the true Pareto front is fully known have been experimented with). The lower the better (a value of 0 indicates that the exact Pareto front has been reached). 

However, and because the true front is known exactly, and is made of a few scattered points (at most 17 for \MULTIZENO9\ in this paper), it is also possible to visually monitor when each point of the front is discovered by the algorithm. This allows some deeper comparison between algorithms even when none has found the whole front. Such {\em attainment plots} will be used in the following, together with more classical plots of hypervolume vs time.

For all experiments, 30 independent runs were performed. Note that all the performance assessment procedures, including the hypervolume calculations, have been achieved using the PISA performance assessment tool suite \cite{Bleuler2003}. 

\paragraph{Stopping Criterion:} Because different fitness evaluations involve different number calls to \YAHSP\ -- and because \YAHSP\ runs can have different computational costs too, depending on the difficulty of the sub-problem being solved -- the stopping criterion was a fixed amount of CPU time rather than the usual number of fitness evaluation. These absolute limits were set to 300, 600, and 900 seconds respectively for  \MULTIZENO3,  \MULTIZENO6, and  \MULTIZENO9.

\section{Experimental Results}
\label{sec:experiments}

\subsection{Comparing Multi-Objective Schemes}
The first series of experiments presented here are concerned with the comparison of the different multi-objective schemes briefly introduced in Section \ref{sec:evolutionaryMOA}.
Figure \ref{fig:zenoHypervolume} displays a summary of experiments of all 4 variants for \MULTIZENO\ instances for both the {\em Cost} and {\em Risk} problems.  

Some clear conclusions can be drawn from these results, that are confirmed by the statistical analyses presented in Table \ref{table:tests} using Wilcoxon signed rank test with 95\% confidence level.
First, looking at the minimal values of the hypervolume reached by the different algorithms shows that, as expected, the difficulty of the problems increases with the number of passengers, and for a given complexity, the {\em Risk} problems are more difficult to solve than the {\em Cost} ones.
Second, from the plots and the statistical tests, it can be seen that NSGA-II is outperformed by all other variants on all problems, SPEA2 by both indicator-based variants on most  instances, and $IBEA_{H^-}$ is a clear winner over $IBEA_{\varepsilon^+}$ except on \MULTIZENO6$_{risk}$.

More precisely, Figure \ref{fig2:zenoParetofront} show the cumulated final populations of all 30 runs in the objective space together with the true Pareto front for \MULTIZENO6-9$_{cost}$ problems: the situation is not as bad as it seemed from Figure \ref{fig:zenoHypervolume}-(e) for \MULTIZENO9$_{cost}$, as most solutions that are returned by $IBEA_{H^-}$ are close to the Pareto front (this is even more true on \MULTIZENO6$_{cost}$ problem).
A dynamic view of the attainment plots is given in Figure \ref{fig:strategiesYahsp}-(c): two points of the Pareto front are more difficult to reach than the others, namely (48,16) and (56,12).

 \begin{figure}[htbp!]
 \centering{
 \subfloat[\MULTIZENO3$_{cost}$]{ \includegraphics[width=0.48\textwidth,height=3cm,bb=50 50 410 302]{zeno3_hyper_{cost}.eps}
 \label{fig-a:zeno3Hypervolume}} 
 \subfloat[\MULTIZENO3$_{risk}$]{\includegraphics[width=0.48\textwidth,height=3cm,bb=50 50 410 302]{zeno3_hyper_{risk}.eps}
\label{fig-b:zeno3Hypervolume}}\\ 
 \subfloat[\MULTIZENO6$_{cost}$]{ \includegraphics[width=0.48\textwidth,height=3cm,bb=50 50 410 302]{zeno6e_hyper_{cost}.eps}
\label{fig-a:zeno6Hypervolume}}  
  \subfloat[\MULTIZENO6$_{risk}$]{ \includegraphics[width=0.48\textwidth,height=3cm,bb=50 50 410 302]{zeno6e_hyper_{risk}.eps}
\label{fig-b:zeno6Hypervolume}}  \\
  \subfloat[\MULTIZENO9$_{cost}$]{\includegraphics[width=0.48\textwidth,height=3cm,bb=50 50 410 302]{zeno9_hyper_{cost}.eps}
\label{fig-a:zeno9Hypervolume}}  
  \subfloat[\MULTIZENO9$_{risk}$]{\includegraphics[width=0.48\textwidth,height=3cm,bb=50 50 410 302]{zeno9_hyper_{risk}.eps}
\label{fig-b:zeno9Hypervolume}}\\
\caption{Evolution of the Hypervolume indicator $I_{H^-}$ (averaged over 30 runs) on \MULTIZENO\ instances (see Table \ref{table:tests} for statistical significances).}
 \label{fig:zenoHypervolume}}
\end{figure}

\begin{figure}[tb]
 \centering{
\subfloat[\MULTIZENO6$_{cost}$]{\includegraphics[width=0.48\textwidth,height=3cm,bb=50 50 410 302]{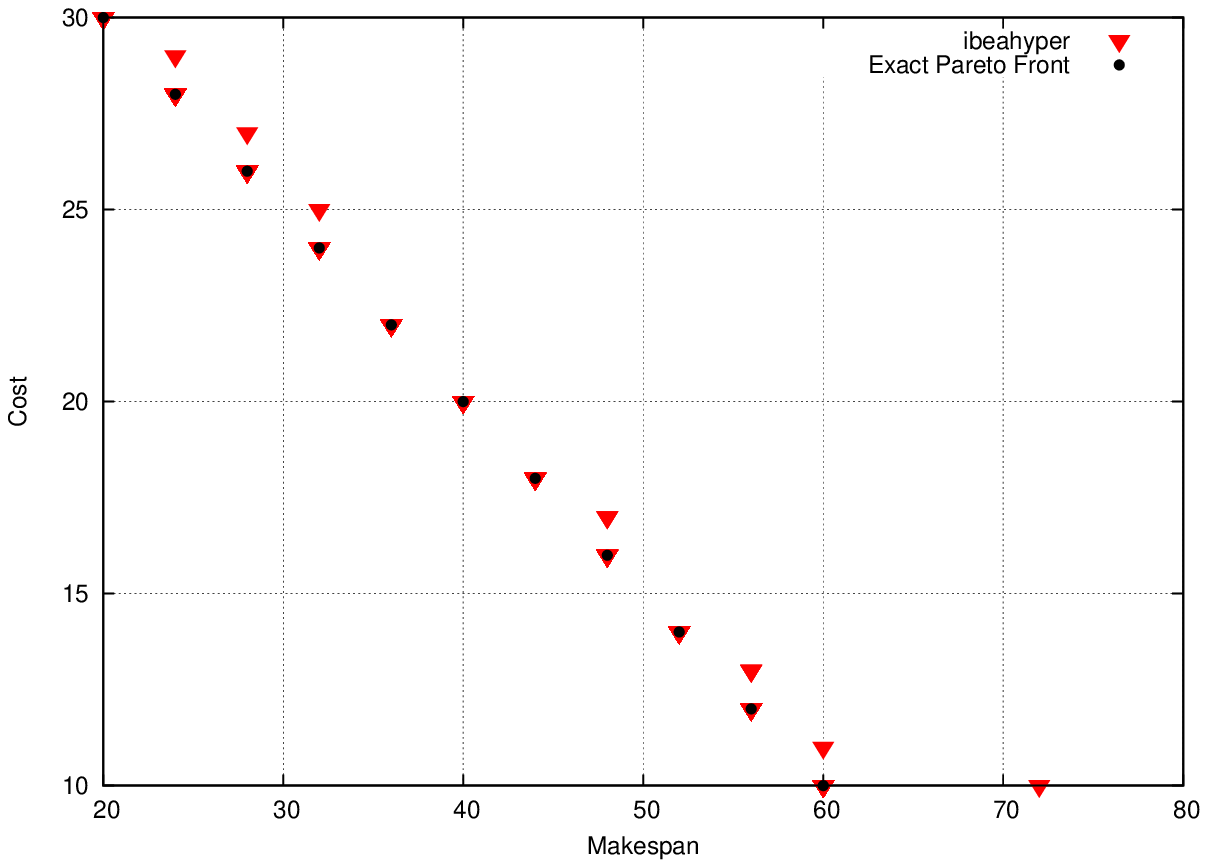}
\label{fig2-b:zeno6AddParetofrontareto}}
 \subfloat[\MULTIZENO9$_{cost}$]{\includegraphics[width=0.48\textwidth,height=3cm,bb=50 50 410 302]{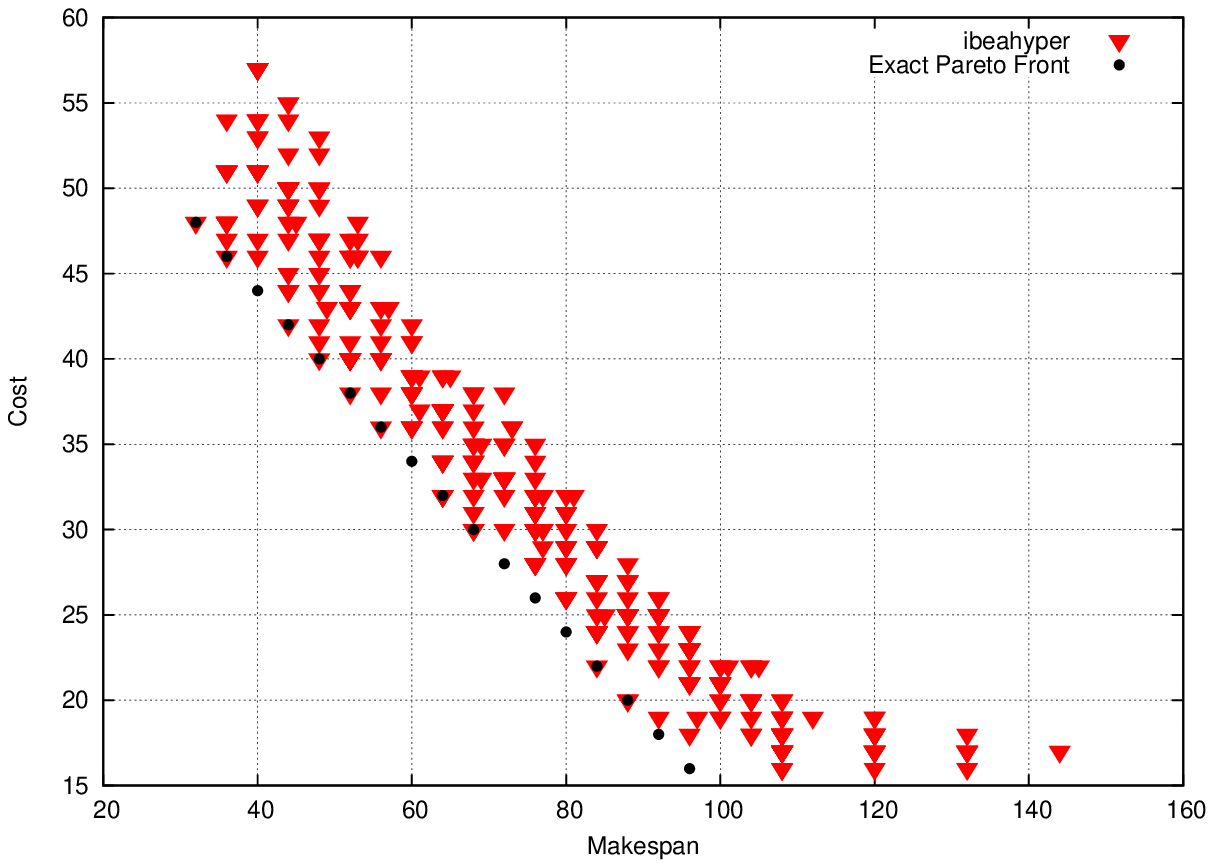}
\label{fig2-c:zeno9AddParetofrontareto}}
 \caption{Pareto fronts of IBEA$_{H^-}$ on \MULTIZENO\ instances.}
 \label{fig2:zenoParetofront}}
\end{figure}

\begin{table*}[tb]
\scriptsize
\caption{Wilcoxon signed rank tests at 95\% confidence level (I$_{H^-}$ metric).}
\label{table:tests}
\centering

\begin{center}
\scriptsize
\begin{tabular}{|l|l|c|c|c|c|}

   \hline
    \multirow{2}*{Instances}  &  \multirow{2}*{Algorithms}	 	&  \multicolumn{4}{c|}{ Algorithms}\\
    \cline{3-6}
			      &            		 	       		& $NSGAII$  &  $IBEA_{\varepsilon^+}$   & $IBEA_{H^-}$  & $SPEA2$  \\
   \hline
  \multirow{4}*{\textit{Zeno3}$_{cost}$} &$NSGAII$       	     &  --     & 		$\equiv$     &  	$\equiv$   	&  	$\equiv$   \\
				
			      &  $IBEA_{\varepsilon^+}$  	   			& $\equiv$  	   & 	--       		& 	$\equiv$ 	&	$\equiv$      \\
			      &    $IBEA_{H^-}$   	 	        	& 	$\equiv$  	&		$\equiv$  &--	&	$\equiv$    \\
			      &    $SPEA2$       		  			& $\equiv$ 		&	$\equiv$ 	&	$\equiv$  			 &  --  \\
  \hline
  \multirow{4}*{\textit{Zeno3}$_{risk}$} & $NSGAII$ 	    			&		-- 					&$\equiv$  		& $\equiv$  	& $\equiv$ \\
	      & $IBEA_{\varepsilon^+}$   	   	 	    		&$\equiv$ 						&-- 			&$\equiv$  	&  \cellcolor [gray]{0.8}$\succ$  \\
	      &  $IBEA_{H^-}$     		&$\equiv$ 			& $\equiv$  						&-- 	 & \cellcolor [gray]{0.8}$\succ$   \\
	      &  $SPEA2$      		&$\equiv$  &  \cellcolor [gray]{0.8}$ \prec$		&  \cellcolor [gray]{0.8}$\prec$  & --   \\
 \hline
  \multirow{4}*{\textit{Zeno6}$_{cost}$} &$NSGAII$       	     &  --     & 		 \cellcolor [gray]{0.8}$ \prec$    &   \cellcolor [gray]{0.8}$ \prec$  	&  	 \cellcolor [gray]{0.8}$ \prec$   \\
				
			      &  $IBEA_{\varepsilon^+}$  	   		&	\cellcolor [gray]{0.8}$\succ$ 	   & 	--       		& 	$\equiv$ 	&	$\equiv$      \\
			      &    $IBEA_{H^-}$   	 	        	& 	\cellcolor [gray]{0.8}$\succ$ 	&		$\equiv$  &--	&	$\equiv$    \\
			      &    $SPEA2$       		  		&	\cellcolor [gray]{0.8}$\succ$ 		&	$\equiv$ 	&	$\equiv$  			 &  --  \\
  \hline

  \multirow{4}*{\textit{Zeno6}$_{risk}$} & $NSGAII$ 	    			&		-- 					& \cellcolor [gray]{0.8}$ \prec$   		&  \cellcolor [gray]{0.8}$ \prec$  	&$\equiv$   \\
	      & $IBEA_{\varepsilon^+}$   	   	 	    		&\cellcolor [gray]{0.8}$\succ$ 						&-- 			&\cellcolor [gray]{0.8}$\succ$  	& \cellcolor [gray]{0.8}$\succ$ \\
	      &  $IBEA_{H^-}$     		&\cellcolor [gray]{0.8}$\succ$ 			&  \cellcolor [gray]{0.8}$ \prec$   						&-- 	 & \cellcolor [gray]{0.8}$\succ$  \\
	      &  $SPEA2$      		& $\equiv$   &  \cellcolor [gray]{0.8}$ \prec$  			& \cellcolor [gray]{0.8}$ \prec$   & --   \\
 \hline
   \hline
  \multirow{4}*{\textit{Zeno9}$_{cost}$} &$NSGAII$       	     &  --     & 		\cellcolor [gray]{0.8}$ \prec$     &  \cellcolor [gray]{0.8}$ \prec$   	&  	\cellcolor [gray]{0.8}$ \prec$  \\
				
			      &  $IBEA_{\varepsilon^+}$  	   			& \cellcolor [gray]{0.8}$\succ$ 	   & 	--       		& 	\cellcolor [gray]{0.8}$ \prec$ 	&	$\equiv$      \\
			      &    $IBEA_{H^-}$   	 	        	& 	\cellcolor [gray]{0.8}$\succ$ 	&		\cellcolor [gray]{0.8}$\succ$   &--	&	$\equiv$    \\
			      &    $SPEA2$       		  			& \cellcolor [gray]{0.8}$\succ$ 		&	$\equiv$ 	&	$\equiv$  			 &  --  \\
  \hline

  \multirow{4}*{\textit{Zeno9}$_{risk}$} &$NSGAII$       	     &  --     & 		\cellcolor [gray]{0.8}$ \prec$     &  	\cellcolor [gray]{0.8}$ \prec$ 	&  	\cellcolor [gray]{0.8}$ \prec$   \\				
			      &  $IBEA_{\varepsilon^+}$  	   		&	\cellcolor [gray]{0.8}$\succ$  	   & 	--     & 	\cellcolor [gray]{0.8}$ \prec$	&	$\equiv$      \\
			    &    $IBEA_{H^-}$     	& \cellcolor [gray]{0.8}$\succ$	& \cellcolor [gray]{0.8}$\succ$	  & 	--	&	$\equiv$     \\
			      &    $SPEA2$       		  			&\cellcolor [gray]{0.8}$\succ$ 	&	$\equiv$ 	&	$\equiv$ 		 &  --  \\
  \hline
\end{tabular} 
\end{center}
\end{table*}

\begin{figure}[tb]
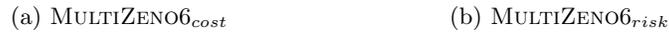

 \centering{

\subfloat[\MULTIZENO6$_{cost}$]{\includegraphics[width=0.48\textwidth,height=3.8cm, bb=50 50 410 302]{zeno6eStratAdd_{cost}:IBEA_{H^-}_{makespan_add-cost}.eps}}
\subfloat[\MULTIZENO6$_{risk}$]{\includegraphics[bb=50 50 410 302,width=0.48\textwidth,height=3.8cm]{zeno6_{risk}:IBEA_{H^-}.eps}}

 \caption{Attainment plots for IBEA$_{H^-}$ on \MULTIZENO6\ instances.}
 \label{fig:attainment}}
\end{figure}

\subsection{Influence of \YAHSP\ Strategy}
\label{resultsStrategies}
Next series of experiments aimed at identifying the influence of the chosen strategy for \YAHSP\ (see Section \ref{sec:strategies}). Figure \ref{fig:strategiesYahsp}-(a) (resp. \ref{fig:strategiesYahsp}-(b)) shows the attainment plots for the strategy in which \YAHSP\ always optimizes the makespan (resp. the cost) on problem \MULTIZENO6$_{cost}$. Both extreme strategies lead to much worse results than the mixed strategy of Figure \ref{fig:attainment}-(a), as no run discovers the whole front (last line, that never leaves the x-axis). Furthermore, and as could be expected, the makespan-only strategy discovers very rapidly the extreme points of the Pareto front that have a small makespan (points (20,30), (24,28) and (28,26)) and hardly discovers the other end of the Pareto front (points with makespan greater than 48), while it is exactly the opposite for the cost-only strategy. This confirms the need for a strategy that incorporates both approaches. 
best possible choice.

Note that similar conclusion could have been drawn from \PARAMILS\ results on parameter tuning (see Section \ref{sec:condition}): the choice of \YAHSP\ strategy was one of the parameters tuned by \PARAMILS\ \ldots and the tuned values for the weights of both strategies were always more or less equal.

\begin{figure*}[h!]
\centering{
\subfloat[\YAHSP\ optimizes makespan]{\includegraphics[width=0.48\textwidth,height=3.4cm,bb=50 50 410 302]{zeno6eStratAdd_{cost}:IBEA_{H^-}_{makespan_add}.eps}}
 \subfloat[\YAHSP\ optimizes cost] {\includegraphics[width=0.48\textwidth,height=3.4cm,bb=50 50 410 302]{zeno6eStratAdd_{cost}:IBEA_{H^-}_{cost}.eps}} \qquad
 \caption{Attainment plots for two search strategies on  \MULTIZENO6$_{cost}$.}
\label{fig:strategiesYahsp}}
\end{figure*}

\subsection{Shape of the Pareto Front}

Figure \ref{fig:allFronts} displays the attainment plots of IBEA$_{H^-}$ for both extreme Pareto fronts shown on Figure \ref{fig:zeno3ParetoFronts} -- while the corresponding plot for the linear case $\alpha=2$ is that of Figure \ref{fig:attainment}-(a). 
Whereas the concave front is fully identified in 40\% of the runs (right), the complete front for the strictly convex case (left) is never reached: in the latter case, the 4 most extreme points are found by 90\% of the runs in less than 200 seconds, while the central points are hardly ever found. We hypothesize that the handling of \YAHSP\ strategy regarding which objective to optimize (see Section \ref{sec:strategies}) has a greater influence in the case of this strictly convex front than when the front is linear ($\alpha=2$) or almost linear, even if strictly concave ($\alpha=2.9$). In any case, no aggregation technique could ever solve the latter case, whereas it is here solved in 40\% of the runs by \DAEYAHSP.

\begin{figure}[tb!]
\centering{
\subfloat[cost(city2)=1.1] {\includegraphics[width=0.48\textwidth,height=3.4cm, bb=50 50 410 302]{./zeno6di_{cost}:IBEA_{H^-}.eps}}
\subfloat[cost(city2)=2.9] {\includegraphics[width=0.48\textwidth,height=3.4cm, bb=50 50 410 302]{./zeno6ds_{cost}:IBEA_{H^-}.eps}}
 \caption{Attainment plots for different Pareto fronts for \MULTIZENO6$_{cost}$.}
\label{fig:allFronts}}
\end{figure}

\section{Conclusion and Perspectives}
\label{sec:conclusion}
The contributions of this paper are twofold. Firstly, \MULTIZENO, an original benchmark test suite for multi-objective temporal planning, has been detailed, and several levers identified that allow to generate more or less complex instances, that have been confirmed experimentally: increasing the number of passengers obviously makes the problem more difficult; modifying the cost of reaching the cities and the duration of the flights is another way to make the problem harder, though deeper work is required to identify the consequences of each modification.
Secondly, several multi-objectivization of \DAEX, an efficient evolutionary planner in the single-objective case, have been proposed.

However, even though the hypervolume-based IBEA$_{H^-}$ clearly emerged as the best choice, the experimental comparison of those variants on the  \MULTIZENO\ benchmark raises more questions than it brings answers. The sparseness of the Pareto Front has been identified as a possible source for the rather poor performance of all variants for moderately large instances, particularly for the {\em risk} type of instances. Some smoothening of the objectives could be beneficial to tackle this issue (e.g., counting for the number of times each risk level is hit rather than simply accounting for the maximal value reached). Another direction of research is to combat the non-symmetry of the results, due to the fact that the embedded planner only optimizes one objective. Further work will investigate a self-adaptive approach to the choice of which objective to give \YAHSP\ to optimize. Finally, the validation of the proposed multi-objective \DAEYAHSP\ can only be complete after a thorough comparison with the existing 
aggregation approaches -- though it is clear that aggregation approaches will not be able to identify the whole Pareto front in case it has some concave parts, whereas the results reported here show that \DAEYAHSP\ can reasonably do it.


{\small
\bibliographystyle{splncs}
\bibliography{emob}
}

\end{document}